# Occlusion Handling in Generic Object Detection: A Review

Kaziwa Saleh
*Doctoral School of Applied Informatics and Applied Mathematics*
*Óbuda University*
*Budapest, Hungary*
kaziwa.saleh@stud.uni-obuda.hu

Sándor Szénási
*John von Neumann Faculty of Informatics*
*Óbuda University*
*Budapest, Hungary*
szenasi.sandor@nik.uni-obuda.hu

Zoltán Vámossy
*John von Neumann Faculty of Informatics*
*Óbuda University*
*Budapest, Hungary*
vamossy.zoltan@nik.uni-obuda.hu

*Abstract*— The significant power of deep learning networks has led to enormous development in object detection. Over the last few years, object detector frameworks have achieved tremendous success in both accuracy and efficiency. However, their ability is far from that of human beings due to several factors, occlusion being one of them. Since occlusion can happen in various locations, scale, and ratio, it is very difficult to handle. In this paper, we address the challenges in occlusion handling in generic object detection in both outdoor and indoor scenes, then we refer to the recent works that have been carried out to overcome these challenges. Finally, we discuss some possible future directions of research.

## I. Introduction

As human beings, we are extremely fast and accurate in detecting and recognizing objects in surrounding environments even under various conditions where the object is partially visible. Our minds are capable of compensating for invisible parts and link the visible areas to identify the object [1]. Computers are still far away from accomplishing this task. However, since the development of deep neural networks and availability of large amount data, there has been a remarkable progress in the field of computer vision in general and object detection in particular. Object detection consists of two sub-tasks: classification and localization of an object.

Object detectors are categorized into two types: one-stage detectors, and two-stage detectors. The latter uses region proposal network to produce regions of interest (ROI) and apply a deep neural network to classify each proposal into class categories. The first type, however, considers the object detection as a regression problem, hence it uses a unified framework to learn the class probabilities and coordinates of bounding boxes. This makes one-stage detectors faster compared to its counterpart. The most effective state-of-art detectors are Faster RCNN [2], SSD [3] and YOLO [4].

However, object detection is a challenging task due to many factors such as clutter, imaging conditions, large number of object categories and instances, and occlusion [5].

Occlusion happens either when the object is hidden by the same type of object, which is called intra-class occlusion, or the object is occluded by a fixed element or an object of another type, this is called inter-class occlusion.

With partial occlusion, the deep neural network based classifiers are less robust compared to humans [6], and it worsens the performance of detectors [7]. Therefore, occlusion handling has been studied extensively such as in pedestrian detection [8] [9] [10] [11], object tracking [12] [13] [14], face detection [15] [16], stereo images [17], car detection [18] [19], semantic part detection [20] [21], etc. However due to a huge number of variations in object category and instances, occlusion handling in generic object detection from a single still image is much harder.

Despite the existence of many datasets for object detection in outdoor scenes, the major problem in dealing with occlusion is the lack of availability of annotated occluded data. Other problems are the detection of occlusion existence, recovering the occluded region(s) of the object, and detecting the occluded object.

Occlusion has also been studied in indoor scenes, but there are several problems which makes it challenging. First, the rigid nature of furniture limits viewing the object fully from different angles. Second, there is not a large-scale dataset of real indoor scenes for occlusion. Finally, the size of objects is relatively smaller compared to those found in outdoors, which means when they are occluded, the visible region may not have sufficient information to be recognized. Thus it becomes harder to regenerate the occluded object.

There are several review and survey papers about object detection available in the literature. Among the most recent ones is the work of Liu et al. in [5] which surveys the available datasets and methods for generic object detection based on deep learning. Also Chen et al. [22] present the challenges and solutions for small object detection, and Zhao et al. [23] provide a review and an analysis for recent deep learning methods for object detection.

Nevertheless, to our best knowledge there is not a recent review about occlusion handling in the literature. Due to the significance of handling occlusion in generic object detection, we provide a review of the recent works that have been done in this field. Then we propose some directions for future research.

This review has been done with respect to object detection in still images. As such, occlusion handling in other applications are out of the scope of this work.

## II. Applications of Object Detection

Two significant applications of object detection in outdoor scenes are autonomous driving [24] [25] and object tracking. In autonomous vehicles, the car needs to have the ability to detect variety of objects on the road such as other vehicles, pedestrians, traffic and road signs, obstacles, etc. [26].

On the other hand, the most apparent application of detecting objects in indoor environments is the 'fetch-and-delivery' task which is the main function of service robots [27]. The robot not only needs to identify the objects around it in order to find the object it is looking for, but it also has to semantically label the region of space it is located in and categorize it, which is another application called scene understanding and categorization [28].

## III. DATASETS

Since the development of deep neural networks, datasets have been the crucial element in the advancement of the field of computer vision. The most widely used datasets for generic object detection are PASCAL VOC [29], MS COCO [30], ImageNet [31], and Open Images [32]. Liu et al. [5] have thoroughly discussed these datasets. To avoid repeating what is already explained in [5], we will focus on the datasets for images in indoor environments.

One of the available datasets is introduced by Ehsani et al. in [33] called DYCE dataset which contains synthetic occluded objects. The images are taken in indoor scenes. There are 11 synthetic scenes, which contain 5 living rooms and 6 kitchen. Every scene has 60 objects and the number of visible (at least 10 visible pixels) objects per image is 17.5.

Furthermore, authors in [34] present TUT indoor dataset which contains 2213 frames containing 4595 object instances from 7 classes. The size of each frame is $1280 \times 720$. But the number of instances are not equal through the classes. There are 1684 maximum instances from a single class while the minimum is 81. The dataset has different backgrounds, light conditions, occlusion, and high inter-class variations. An object detector (Faster RCNN using ResNet-101 [35]) is trained to generate proposal annotations for half of the dataset. While the other smaller half is manually annotated by a human. Thus a faster bounding box annotation is presented in the study.

To efficiently train a model on occlusions in real indoor scenes we need a large scale fully annotated dataset with ground truth labels for occluded objects in the scene. Unfortunately, there is no such dataset yet.

For outdoor scenes, Qi et al. [36] created KINS dataset from KITTI [37] for amodal instance segmentation. It contains amodal instance segmentation masks and corresponding occlusion order. Three expert annotators marked every image. Ambiguities were resolved by crowdsourcing to ensure that the occluded regions were labeled consistently.

Apart from the previously mentioned datasets, there are others such as ScanObjectNN (contains indoor scenes created with CAD data) [38], BigBird [39], NYC v2 [40], Places [41], and datasets for 3D indoor reconstruction and Simultaneous localization and mapping (SLAM) mentioned in [42]. But those are out of the scope of this review.

## IV. OCCLUSION HANDLING IN OUTDOOR SCENES

In this section we address the challenges in occlusion handling in outdoor environments and the solutions for each of them.

### A. Data Collection

Because of large variety of object categories and instances, collecting and labeling a dataset with possible occlusions of each instance in every category seems impossible. Consequently, many studies rely on synthetic datasets or automatically generated examples:

*1) Generative Adversarial Networks (GAN)*: Since their inventions by Goodfellow et al. [43], Generative Adversarial Networks (GAN) have been studied widely to train a generative model. The framework has two adversarial networks which are trained simultaneously, namely a generator and a discriminator. Whereas the generator is trained to learn mapping samples from a random latent space to data, the discriminator is trained to differentiate between real and generated (fake) samples. The goal of the generator is to fool the discriminator through making samples that appear similar to the real data as possible. To ensure that the generated data comes from a particular class or data, Conditional Generative Adversarial Networks (cGAN) can be used [44].

Wang et al. [45] argue that potential occlusions and deformations cannot be covered even by large scale datasets. Thus they propose a strategy that uses two adversarial networks (shown in "Fig. 1") to generate examples from COCO dataset that would be hard for a Fast-RCNN to classify. The first adversarial network is Adversarial Spatial Dropout Network (ASDN) which learns how to occlude an object, and the second network is Adversarial Spatial Transformer Network (ASTN) which learns how to rotate object parts to create deformations. By training these two networks simultaneously against the Fast-RCNN, the latter learns to deal with occlusions and deformations.

*2) Amodal perception:* is the ability of inferring the object's physical structure when parts of it are occluded. Recent studies such as [46] [47] [36] have used it for instance segmentation. Amodal segmentation is significant for occlusion handling because through comparing amodal and modal segmentation masks the existence, degree, contour, and parts of occlusion can be deduced [46].

Nonetheless, the difficulty of data preparation for amodal instance segmentation makes it a challenging task. Therefore, Li and Malik [46] create amodal training data by adding synthetic occlusion to the modal mask. First, an image patch that has at least one foreground object instance is randomly cropped. Afterwards, the patch it is overlaid by random object instances that are extracted from other images. The random position and scale of the overlaid object ensures a mild occlusion with the bottom patch. Then by using the original modal segmentation mask, the pixels congruous to the mask in each patch are labeled as positive (belongs to the object), negative (background), and unknown (belongs to other objects). Eventually, the original modal mask contains parts of the occluded object in the composite image that were originally visible. This becomes the true amodal segmentation mask for the composite patch.

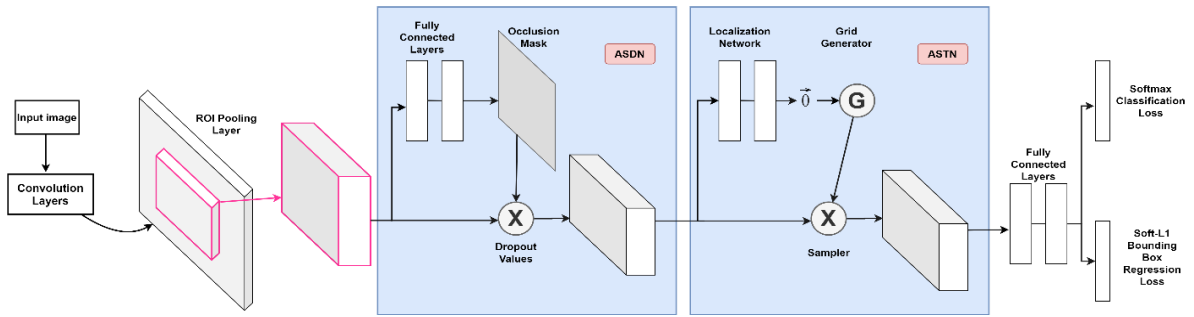

Fig. 1. Proposed architecture for combining ASDN and ASTN in [45]

The generated composite image and the amodal mask are used to train a CNN to attain the original mask. When tested on images with real occlusion, the model can effectively predict amodal masks even if it is trained on synthetic data.

### B. Occlusion Detection

One of key problems in occlusion handling is to determine whether the object in question is occluded by other objects or not. It is hard to tell if the observed appearance of the object in the image is the object's real shape or it is a result of occlusion.

Qi et al. in [36] propose Multi-Level Coding (MLC) network with an occlusion classification branch to improve the amodal perception ability of inferring the occluded parts. MLC has two branches: extraction and combination. In the first branch the abstract global features of the objects are extracted. To produce the masks for visible and invisible parts, the combination branch fuses global and specific local features. Meanwhile, the occlusion classification predicts the existence of occlusion which enhances the network's amodal perception. Experimental results on KINS dataset show that amodal and inmodal instance segmentation can be enhanced through using the proposed model.

### C. Generating the Occluded Regions

A major challenge in handling occlusion is determining how the invisible parts of the object can be recovered. Currently there are three solutions:

*1) Amodal instance segmentation:* In addition to previously metioned works in amodal instance segmenation, Follmann et al. [48] propose an end-to-end trainable amodal instance segmentation model dubbed Occlusion R-CNN (ORCNN). The model is an extension of Mask R-CNN with amodal mask head and an occlusion mask head for predicting amodal, inmodal, and occlusion masks for object instances concurrently in a single forward pass (as seen in "Fig. 2"). The authors also introduce a new D2S amodal dataset and COCOA cls. The first one is based on D2S [49] and the latter is originated from COCOA dataset from [47]. Data augmentation is used to include moderate to heavy occlusion occluded objects in D2S amodal dataset. The results verify that the model obtains competitive results on D2S amodal without any amodally annotated data, it even performs better than its baselines on COCOA cls dataset.

*2) Partial completion:* a self-supervised framework developed by Zhan et al. [50] for partial completion of occluded objects for scene de-occlusion. The framework depends on two principles of partial completion concept. First, in case of having an object occluded by several other objects, the partial completion can be performed progressively with one object involved at a time. Second, by intentionally pruning an occluded object and training a network to recreate the unpruned object, the network can learn to partially complete the occluded object. The authors attain partial completion through two networks: Partial Completion Network-mask (PCNet-M) and Partial Completion Network-content (PCNet-C). While the first network is used to generate the occlusion mask of the occluded object corresponding to the occluding object, the second network furnishes the mask with RGB content. The framework is tested on KINS [36] dataset and COCOA [47], and the results illustrate a similar performance to fully supervised baselines although the framework is trained without ground truth occlusion ordering and amodal masks.

*3) Context Encoders:* Pathak et al. [51] suggest a convolutional neural network (CNN) that is able to generate the missing patches of an image based on context. The proposed model has an encoder and a decdoer. While the encoder produces a compact latent feature representation from the image's context, the decoder can generate the missing part of the image from the produced representation. Due to its unsupervised nature, the model has to learn the semantic of the image and generate a reasonable hypothesis for the absent parts. Therefore, the model is trained to reduce a reconstruction loss which catches the structure of the absent part based on the context, and an adversarial loss which selects a specific mode from the distribution. The results show that the model can inpaint sematic parts of an image better than its baselines. However, the performance of the model deteriorates with high-level textures regions.

### D. Occluded Object Detection

The accuracy of state-of-art object detectors and deep neural network based classifiers degrade with partial occlusion [52] [7]. As such, many studies focus on improving the ability of current methods to recognize and localize occluded objects. We divide these into two categories:

*1) Classification:* According to Fawzi and Frossard [53] Deep Convolution Neural Networks (DCNN) are not robust under partial occlusion.

DeVries and Taylor [54] argue that CNNs are prone to overfitting which leads to poor generalization in case of occlusion. Thus they propose a simple regularization technique for CNN called Cutout, where they augment the training data with partially occluded images. For each input image, a pixel is randomly chosen as a center point for a fixed-size zero mask to drop out neighboring section of the image. However, increasing the training data leads to the increase in training time and cost [7].

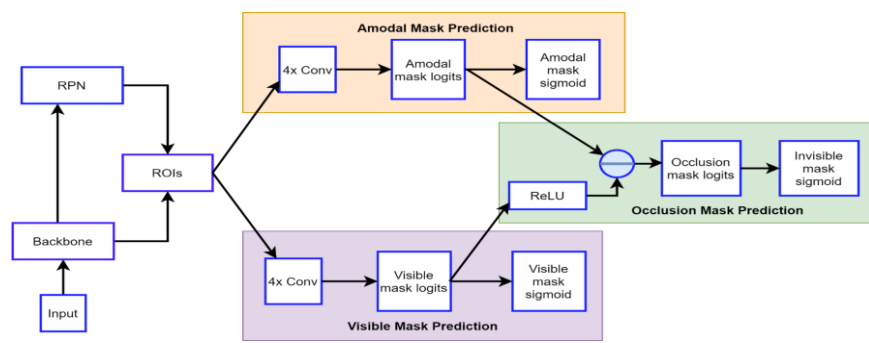

Fig. 2. Architecture of ORCNN [48]

On the other hand, Xiao et al. [55] propose TDAPNet to address two problems in DCNN: overfitting, and contamination of occlusion during feature extraction. They propose TDAPNet, a deep network which has three parts: prototype learning, partial matching, and a top-down attention mechanism. The first two parts help to solve the first addressed issue, whereas the third part deals with the second issue. After extracting the features by DCNN, prototype learning is applied. Then, partial matching compares features and prototypes by only comparing the visible parts thus removing irrelevant feature vectors. Finally, the top-down attention regulation produces more pure features around the occluder by filtering out irregular activations caused by occlusion. Through removing occluded features in the lower layers, the model becomes more robust against occlusion. But based on the experiments done by Kortylewski et al. [7] on the images with real occlusion, the model is not as reliable as it is with artificial occlusion.

Moreover, many recent studies have focused on using compositional models to handle occlusions [52] [56] [57]. Through compositional models, an object can be represented in terms of its parts and their spatial structure [52]. Compositional models have two benefits: 1) we can create an occlusion model which ignores regions of the model that are occluded. 2) The model can provide explanation about its results such as: the location of object's detected individual parts, and the occluded regions of the object [52].

Kortylewski et al. in [52] combine the discriminative power of DCNN and the ability of compositional to generalize well to partially occluded objects. In the proposed model, a standard DCNN is used for image classification during training. The extracted features are grouped into dictionaries. The elements within the dictionaries are similar to object part detectors and learn the spatial allocation of parts of each class. During testing, the DCNN tries to classify the input image through a feedforward pass. It is probable that the image is partially occluded if the network cannot certainly predict its class. Then the extracted features are used to detect the parts of the compositional model. The results demonstrate that the model can recognize partially occluded 3D objects better than DCNN even when it has been trained on occlusion-free data. However, the model is not as discriminative as DCNN at recognizing non-occluded objects [57].

*2) Detection:* despite the success of state-of-art detectors in detecting non-occluded objects, detection of occluded objects remains an open problem. Nevertheless, several recent works have implemented compositional models to detect partially occluded objects.

Kortylewski et al. in [57] introduce a unified model of DCNN and compositional models called Compositional Convolutional Neural Network (CompositionalNet). In their proposed architecture, the authors apply a differentiable generative compositional layer instead of the fully-connected head in DCNN. The network can robustly classify the partially occluded objects and localize the occluders due to the generative characteristic of the compositional layer. The experiments show that, although CompositionalNet is trained with class labels only, it can localize occluders correctly. It also outperforms a standard DCNN and other related techniques in classifying partially occluded objects despite not being trained on occluded objects.

However, according to Wang et al. [56] CompositionalNets have no explicit separate representation of context from the object, subsequently the context has a negative effect on detection in severe occlusion. And there is no mechanism for robust bounding box prediction for partially occluded objects in CompositionalNets. To manage the effect of context on detecting the occluded objects, the authors suggest using bounding box annotations for segmenting the context. Meanwhile, the part-based voting technique in CompositionalNets is extended to consider the vote for the two opposite corners of the bounding box beside the object center. The results demonstrate that the proposed context-aware CompositionalNet can robustly detect and estimate bounding boxes even under severe occlusion.

Apart from the previously mentioned efforts, contextual information [58] [59] [60], and deformable convolution [61] [62] [63] can be used to reduce occlusion.

V. OCCLUSION HANDLING IN INDOOR SCENES

Due to complex and diverse layout of indoor scene structure, clutter, occlusion, and illumination, object detection is more challenging compared to that of outdoor scenes [42]. In this section we present the challenges of object occlusion handling in indoor environments, and existent methods in the literature to solve them:

*A. Scene Structure*

The layout and design of indoor elements and furniture, prevents obtaining full view of objects or even partial view in some cases. For example, looking for a hidden box behind other objects inside a cupboard. Because of the static and rigid nature of the cupboard, it would be impossible to view the object from different viewpoints. To overcome this issue, there are two solutions:

*1) Turntable:* is used to avoid occlusion by placing the objects on a round table and allow the camera that is installed on a robot to go around the table to capture the objects in different viewpoints [64]. The different views are used to provide the next-best view (NBV) of the object to model the invisible parts.

*2) Interactive manipulation:* authors of [65] argue that finding an occluded object within a cluttered scene cannot be done by using a single image. Therefore using a robotic arm with wrist-mounted RGB and depth camera, active perception and interactive perception are applied to find the object of interest. While active perception is where the camera is moved to capture the object from several viewpoints, interactive perception provides better understanding of the scene from interactions. Then a reinforcement learning based control algorithm and a color detector are used to find target object of a particular color.

Although the agent is trained with a detached gripper during simulation, some end-effector poses are not kinematically feasible in real world.

Also Dogar et al [66] state that the problem of searching for an object requires both perception and manipulation to move the objects that might hide the targeted object. They propose two search algorithms, the greedy search algorithm that considers visibility and accessibility relations between objects; and the connected components algorithm which utilizes the expected time to find the target as an optimization criterion.

Interactive manipulation of an object by a robot arm in front of a static camera is used by Krainin et al. [67] to generate a 3D model of an object.

Although the mentioned techniques might be necessary and required in certain cases where applicable, they require a controlled setup of the environment which is unrealistic and not always possible in real world.

*B. Training Data*

There is not a large-scale database available for object occlusion for indoor scenes.

Georgakis et al [27] argue that creating annotated dataset that covers all the possibilities of indoor environments such as differences in viewpoint, lighting conditions, occlusion and clutter, would be laborious and time consuming. Meanwhile the trained models poorly generalize across various environments and backgrounds. Therefore, they propose an approach to generate a new dataset from two other datasets: GMU-Kitchens and Washington RGB-D Scenes v2. The synthesized images are generated by superimposing object instances at several positions, scales, pose, and location.

On the other hand, Dwibedi et al [68] suggest an easy cut and paste method to synthesize training data with minimal effort. Their key intuition is based on state-of-the-art detectors such as Faster RCNN that mostly work on local region-based features rather than global-based features. The method automatically cuts object instances and pastes them on random backgrounds. However, to avoid subtle pixel artifacts the training algorithms is forced to ignore the artifacts and concentrate on the object appearance instead.

*C. Recover the Occluded Regions*

Occlusion can happen in different scale, location, and levels. Thus it is hard to train a model to regenerate the invisible regions of an object. The following methods are used to segment and regenerate the occluded area of objects:

*1) GAN:* Ehsani et al. [33] use GAN to address the problem of generating the occluded region of objects. Their proposed model called SeGAN segments the invisible part first, then produces its appearance by painting. Based on an input image and a segmentation of the visible region of the object, the model produces an RGB image where the invisible region of the object is rebuilt. To do this, the model consists of two parts: the segmentation part which is a CNN that outputs a mask for the object by using the information from visible regions, and a painting part which uses a cGAN to produce the occluded parts of the object. The loss function of the SeGAN model is a combination of the losses for segmentation and painting. "Fig. 3", shows the architecture of SeGAN model. The authors report better results of SeGAN compared to its baselines on their photo-realistic DYCE dataset.

*2) Semantic segmentation:* Purkait et al. [69] propose that instead of assigning a label per pixel in an image, a set of semantic labels can be achieved indicating visibility or invisibility per each pixel. Their work uses a synthesized dataset that was augmented from SUNCG dataset to predict the semantic category of visible regions and occluded regions. They use a U-Net architecture [70] that has an encoder and a decoder with ReLUs, but in the final layer a sigmoid activation function is used with a group-wise softmax. There are skip connections from the encoder to the decoder.

The results of the network with cross-entropy loss is compared with the network that utilizes a grouped semantic loss, the latter produces better results in predicting semantic labels for the occluded pixels. However since the training set is synthetic, the implementation of the proposed method is bounded to environments where such a synthetic models are found. In real world scenes, certain estimation of occluded parts has to be considered.

*3) Instance segmentation:* By extending the Mask RCNN architecture, Wada et al. [71] propose a model for occlusion segmentation which they call it "relook Mask RCNN". To train their model, the authors create a dataset of synthesized images of a pile of object created from the instances of the objects. Instead of considering instance occlusion segmentation as a single class (visible only) problem, the authors look at it as a multi-class (visible, occluded). They also consider the relationship between the masks to infer the occlusion status of the instances correctly. To learn the relationship, the instance masks predicted from the Mask RCNN is converted to a density map to predict the instance mask in the second stage (relook stage). As an output, the model predicts three masks: visible, occluded, and other (not belonging to the object).

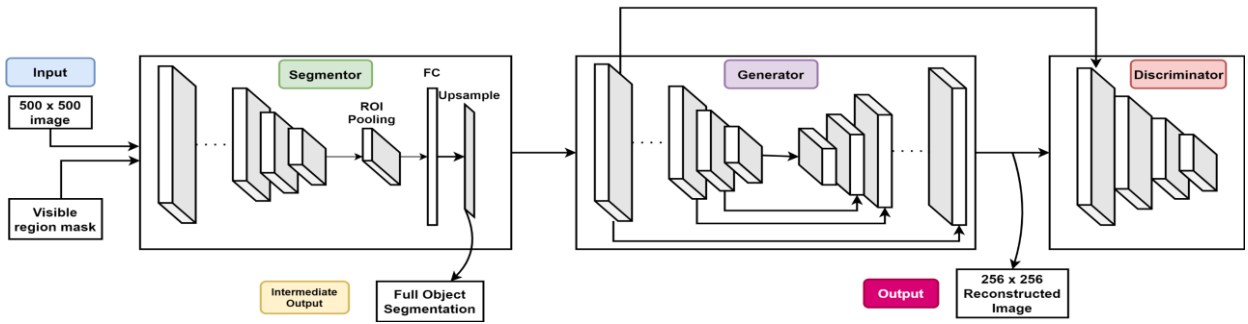

Fig. 3. Architecture of SeGAN as proposed in [33]

The proposed system is tested on real picking task from a stack of objects and the reported results illustrate the effectiveness of the system. However the system requires a dataset with all possibilities of occlusion state of the objects and their corresponding labels and masks, the effort for achieving this increases exponentially with the increase of the number of objects.

*4) Contextual information:* Hueting et al. [72] introduce SEETHROUGH model that utilizes the fact that there is a considerable regularity of objects co-occurrence in indoor scenes. By considering that co-occurrence information as explicit priors, the object identity, location and orientation can be predicted even with heavy inter- or intra-class occlusions. The authors train a neural network on real indoor annotated images to extract 2D keypoints. The extracted key points are fed into a 3D candidate object generation stage. Then object co-occurrence statistics extracted from a large 3D scene database is used to solve a selection problem between 3D object proposals. The process is repeated to use the location of the already discovered objects for incrementally detecting nearby candidates with low keypoint response. The results show that SeeThrough can more accurately detect chairs in scenes even with medium or heavy occlusion than its two baselines, Faster-3D RCNN and SeeingChairs.

## VI. Discussion and Future Directions

Based on the reported results of GAN, amodal segmentation, and compositional models, we see a great potential of these techniques in occlusion detection and handling. GAN have been effectively used to enhance features when there are not enough of them in small object detection [73], and to regenerate occluded faces [74].

However, there remains a set of key problems in occlusion handling that could serve as a future direction of research in the field.

First, there is no large-scale dataset that could be used for object occlusion in indoor environments. The available ones currently are mostly synthetic datasets, which means when a model is trained on such a dataset it might not generalize well for a real world scene. In addition, the available datasets for outdoor scenes lack sufficient annotations for occluded regions.

Second, in order to apply de-occlusion on an object, it is vital that we first determine that the object is occluded or not. Although amodal segmentation can be used to resolve this issue, this solution may be ineffective if the annotation is incorrect or insufficient.

Finally, increase the ability of current models to regenerate the occluded objects under different degrees of occlusion using the real world training data instead of synthetic data.

## VII. Conclusion

Object detection has come a long way since the invention of region based networks. Not only they are more accurate than before but they also achieve real-time results. Although there is abundant studies in generic and specific object category detection, occlusion handling in generic object detection remains comparatively unexplored. Many prominent models fail to detect objects when they are occluded. In this review, we have presented the most recent works that endeavor to resolve occlusions and in many cases regenerate the object. We have also addressed the future directions for research if we want to overcome challenges of occlusion handling in generic object detection.

## Acknowledgement

The authors would like to thank the GPGPU Programming Research Group of Óbuda University for its valuable support.